# AI Literacy in Low-Resource Languages: Insights from creating AI in Yoruba videos

**Wuraola Oyewusi**
Tech in Yoruba
oyewusiwuraola@gmail.com

## Abstract

To effectively navigate the AI revolution, AI literacy is crucial. However, content predominantly exists in dominant languages, creating a gap for low-resource languages like Yoruba (41 million native speakers). This case study explores bridging this gap by creating and distributing AI videos in Yoruba.The project developed 26 videos covering foundational, intermediate, and advanced AI concepts, leveraging storytelling and accessible explanations. These videos were created using a cost-effective methodology and distributed across YouTube, LinkedIn, and Twitter, reaching an estimated global audience of 22 countries. Analysis of YouTube reveals insights into viewing patterns, with the 25-44 age group contributing the most views. Notably, over half of the traffic originated from external sources, highlighting the potential of cross-platform promotion.This study demonstrates the feasibility and impact of creating AI literacy content in low-resource languages. It emphasizes that accurate interpretation requires both technical expertise in AI and fluency in the target language. This work contributes a replicable methodology, a 22-word Yoruba AI vocabulary, and data-driven insights into audience demographics and acquisition channel

## 1 Introduction

### 1.1 The Rise of AI and the Need for Literacy

With the explosion of data, computing power, and learning algorithms, Artificial Intelligence (AI) has unlocked possibilities once thought impossible. AI is already shaping our experiences with technology, both directly and indirectly, and its influence is growing rapidly. This rapid evolution has sparked a shift in how we approach AI literacy, moving beyond a limited scope in higher education to include early childhood education, preparing a generation equipped to understand this transformative technology.

### 1.2 Unequal Access: The Language Gap in AI Literacy

AI literacy refers to the understanding and knowledge of AI concepts, technologies, and their implications on society. It's about being able to comprehend how AI systems work,their limitations,and their potential impact on various aspects of our lives(Long et al., 2023).AI literacy is a critical skill for everyone, regardless of language or background. While there's a global need for AI literacy, its delivery faces an unequal distribution based on language resources,there is a concerning gap in AI literacy education, out of the more than 7000 Languages spoken worldwide, English dominates the internet Ta & Lee (2023), most AI educational content, both formal and informal, exists in English, creating a gap for low-resource languages.

### 1.3 Efforts Towards Inclusion: Non-English Initiatives in AI Literacy

This issue is not entirely unaddressed. Other high-resource languages besides English are starting to see efforts in AI literacy localization. For example, as of 2021, 11 countries (Armenia, Austria,Belgium,China,India,Republic of Korea ,Kuwait, Portugal, Qatar,Serbia and UAE) with non-English official languages have implemented government-endorsed AI curricula in one or all of primary, middle, and high school levels (UNESCO, 2022). Furthermore, other studies have explored the design and implementation of AI education and literacy in non-English contexts. One study by Gibellini et al. (2023)examined how AI education designed for middle schools in disadvantaged urban or rural areas of four European countries impacted inclusion and diversity. While they mention various digital tools being developed in English, the specific languages of delivery remain unclear. However, the study's focus on non-English countries like Greece, Bulgaria, Romania, and Italy suggests the use of non-English languages. Another study by Mi & Ok (2022) analyzed the existing curriculum for AI literacy in South Korea's software universities, specifically for non-computer science majors. This initiative aims to improve the quality and future potential of talents to be digital and AI literate.

### 1.4 Africa's Linguistic Landscape: A Call for Inclusive AI Education

While limited research exists on AI literacy curricula specifically for African contexts,the call for its inclusion is growing. Scholars like eOyelere et al. (2022) highlight the need for contextualized AI resources in African elementary schools, and Onyejegbu (2023) emphasizes the growing need for including AI ethics, a core aspect of AI literacy in Higher Education curricula across West Africa .Notably, existing studies don't explicitly mention using African languages for delivering AI education. This gap is crucial to address, considering Africa's linguistic diversity. Over 2,000 languages are spoken by a billion people across the continent, yet most are classified as "low-resource" due to a lack of data, digital tools, and research expertise compared to dominant languages (Ògúnrẹ̀mí et al., 2023) .This creates a significant barrier to AI literacy education for these communities.Therefore, broader and more inclusive AI literacy education is essential, particularly for low-resource languages.

### 1.5 Bridging the Gap: Exploring AI Literacy in Yoruba

Broader and more inclusive AI literacy education is essential, particularly for low-resource languages. This work draws insights from developing an AI video series in Yoruba, a low-resource language with over 41 million native speakers primarily in Southwestern Nigeria and Benin Republic(Worlddata.info, 2021). It explores the process of creating AI literacy resources for such languages, aiming to bridge the gap and empower communities with the knowledge and understanding of this transformative technology.

## 2 Method

The development and production for AI in Yoruba videos follow three key steps : Ideation,Production and Distribution as shown in Figure 1.

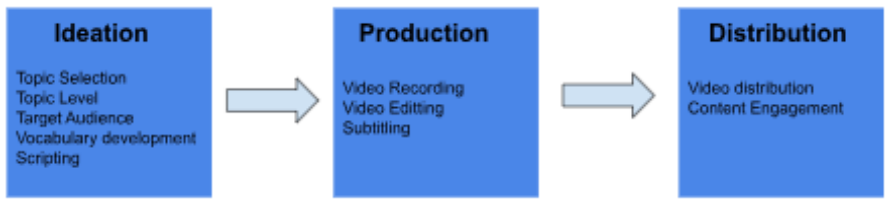

Figure 1: Process for creating AI in Yoruba Videos

*Ideation:* At this stage, the topic, topic level(Foundational,Intermediate and Advanced), and target audience(technical and non-technical) are determined. All videos are designed to cater to a broad range of viewers,Foundational builds basic AI understanding, Intermediate dives deeper into specific AI techniques and Advanced tackles more complex concepts. The chosen engagement styles are storytelling and closest-analogy explainers.

The scripting approach focuses on emphasizing key discussion points rather than providing full-fledged narratives. A crucial aspect, particularly when creating technical content in low-resource languages, is developing suitable vocabulary that may not previously exist in the language. The goal is to be succinct and precise while ensuring clear understanding

*Production:* This project prioritized cost-effectiveness by leveraging readily available tools and minimizing personnel and space needs. While three of the twenty-six videos utilized a volunteer videographer and camera setup, the remaining videos relied on accessible tools like laptop webcams, free video software and smartphones, enabling single-person production. Editing was kept minimal, focusing solely on adding title slides and English subtitles using native Windows video editing software. On average, each video is five minutes long, with an estimated 45 minutes spent subtitling each one.

*Distribution:* Each video was uploaded to three platforms: YouTube, LinkedIn, and X(Twitter). Short notes accompanied each upload, highlighting the video title, including any new vocabulary where applicable, and using relevant hashtags that would resonate with the target audience. Additionally, LinkedIn and X(Twitter) posts included the YouTube video link. This strategy aimed to improve YouTube engagement and accommodate Twitter's 2-minute, 20-second video limit

## 3 RESULTS

To illustrate the range of topics covered and their suitability for diverse audiences, Table 1 presents the titles and corresponding topic levels (Foundational, Intermediate, Advanced) of the 26 AI in Yoruba videos created between May 2022 and August 2023.

Table 1: AI in Yoruba Video Titles and Topic Level

| TOPIC LEVEL | VIDEO TITLES |
| --- | --- |
| Foundational | Introduction to AI in Yoruba (Part 1) |
| | Introduction to AI in Yoruba (Part 2) |
| | Introduction to AI in Yoruba (Part 3) |
| | Introduction to Computer Vision in Yoruba |
| | Introduction to Data Science in Yoruba |
| | Machine Translation, explained in Yoruba |
| | Recommender Systems, explained in Yoruba |
| | Voice Assistants explained in Yoruba |
| | ChatGPT explained in Yoruba |
| Intermediate | AI for Everyone: Types of Machine Learning in Yoruba |
| | AI for Everyone: Self Driving Cars in Yoruba |
| | AI for Everyone: Topic Modelling in Yoruba |
| | AI for Everyone: Clustering Algorithms in Yoruba |
| | AI for Everyone: Machine Learning for Loan Prediction in Yoruba |
| | AI for Everyone: ML for Sentiment analysis of Customer reviews in Yoruba |
| | AI for Everyone: Machine Learning for Facial detection and recognition in Yoruba |
| Advanced | AI for Everyone: Graphical Processing Units (GPUs) in Yoruba |
| | AI for Everyone: Artificial Neural Networks in Yoruba |
| | AI for Everyone: Transfer Learning in Yoruba |
| | AI for Everyone: Generative Adversarial Networks (GANs) in Yoruba |
| | AI for Everyone: Generative AI in Yoruba |
| | AI for Everyone: Embeddings for Text Representation in Yoruba |
| | AI for Everyone: Large Language Models (LLMs) in Yoruba |
| | AI for Everyone: Semantic Similarity in Yoruba |
| | AI for Everyone: Named Entity Recognition in Yoruba |
| | AI for Everyone: Optical Character Recognition (OCR) in Yoruba |

One of the key challenges in creating technical content for low-resource languages like Yoruba is the lack of existing vocabulary for certain concepts. While direct translation might seem like a straightforward solution, it's not solely a matter of language proficiency. Understanding how AI works is crucial for accurately interpreting these concepts and conveying them effectively. Therefore, the ideal individuals for this task are technical experts who are also speakers of the target language. Table 2 presents the new vocabulary terms created during the development of AI in Yoruba videos, along with their literal meanings. These newly developed terms contribute significantly to making AI concepts more accessible in Yoruba Language

Table 2: Yoruba words for concepts in AI in Yoruba videos

| AI Concept | Yoruba Word | Literal Meaning |
| --- | --- | --- |
| Artificial Intelligence(AI) | Ọgbọ́n Àpinlẹ̀rọ | Ọgbọ́n : Intelligence/Wisdom, Àpinlẹ̀rọ : A version made from scratch or locally made, it's typically used to connote some form of substitution. |
| Machine Learning | Ìkẹ́ẹ̀kọ́ èrọ | Ìkẹ́ẹ̀kọ́ : Learning, Ẹrọ: Machine |
| Computer Vision | Ìríran èrọ | Ìríran : To see, Ẹrọ: Machine |
| Data Science | Sáyẹnsì àkíyèsí | Sáyẹnsì : Science, Àkíyèsí: Observation |
| Machine Translation | Òngbifọ̀ èrọ | Òngbifọ̀ : Translation, Ẹrọ: Machine |
| Recommender Systems | Ẹrọ Àsàyàn | Ẹrọ : Machine, Àsàyàn : Preference |
| Voice Assistants | Ohùn aranilọ́wọ́ | Ohùn : Voice, Aranilọ́wọ́: Helper |
| Clustering Algorithms | Algọ́rídímù Ẹlẹ́gbẹ́jẹgbẹ́ | Algọ́rídímù : Algorithm, Ẹlẹ́gbẹ́jẹgbẹ́ : Peer to Peer |
| Sentiment Analysis | Àgbéyẹ̀wò Ìtara | Àgbéyẹ̀wò : Analysis, Ìtara : Sentiment |
| Topic Modelling | Ìsọ̀rí àkọ́lé | Ìsọ̀rí : category, Àkọ́lé : Topic |
| Large Language Models | Àwòṣe èdè nlá | Àwòṣe èdè : Language Model, Nlá : Big |
| Chatbots | Asọ̀rọ̀gbèsì | Asọ̀rọ̀gbèsì : Responsive conversation tool |
| Transformers | Akẹ́ẹ̀kọ́gbésìjáde | Akẹ́ẹ̀kọ́gbésìjáde : A learner that can produce results |
| Self attention | Ìfojúsíaraẹni | Ìfojúsíaraẹni : Paying attention to oneself |
| Positional encoding | Ìṣààmìsí ipò | Ìṣààmìsí : Making a mark, Ipò : Position |
| Named Entity Recognition | Afinipeni | Afinipeni : Calling one as it is |
| Semantic Similarity | Ìjọra ọ̀rọ̀ atúnmọ̀ | Ìjọra ọ̀rọ̀ : word similarity, Atúnmọ̀ : Semantic |
| Text Embedding | Aṣojú ọ̀rọ̀ alásopọ̀ | Aṣojú ọ̀rọ̀ : Word representation, Alásopọ̀ : Embedded or Connected |
| Search Engine | Ọpọ́n àwárí | Ọpọ́n : Interface, Àwárí : Finding |
| Generative AI | Ọgbọ́n àpinlẹ̀rọ aṣèdá | Ọgbọ́n àpinlẹ̀rọ : AI, aṣèdá : Creator or Generator |
| Generative Adversarial Networks(GANs) | Algọ́rídímù aṣèdá | Algọ́rídímù : Algorithm, Aṣèdá : Creator or Generator |
| Optical Character Recognition(OCR) | Asawòràndikíkọ | Asawòràndikíkọ : To make an image into written form |

## 4 Discussion

### 4.1 Data and Analysis Considerations

This work analyzed data from three social media platforms: LinkedIn, X(Twitter), and YouTube. While the choice of distribution platform was motivated by personal convenience, each platform offers valuable insights. It is important to acknowledge that they prioritize different metrics, and YouTube provides the most granular data for in-depth analysis. Three metrics are common across all platforms: number of impressions, comments, and reactions. Impressions refer to the number of times a video is displayed on a user's screen. It's a measure for visibility and not necessarily engagement but a good proxy for reach. Figure 2 shows the count and proportion of impressions across platforms in descending order.While this provides a general overview of the distribution and reach of AI in Yoruba videos, it's important to acknowledge limitations in the data. First, the videos were uploaded between May 2022 and August 2023, and the data was extracted in February 2024. This time difference can introduce biases and affect the observed distribution.My experience suggests most impressions occur within 3 days of posting. However, all videos analyzed are at least 5 months old, potentially mitigating the possible biases by having established reach and engagement.Second, factors like existing followers, activity on each platform, and their unique distribution algorithms can significantly influence the reach of content. I didn't actively engage on YouTube before the series, but creating the video series increased follower engagement on all platforms.These limitations should be considered when interpreting the findings to avoid drawing inaccurate conclusions

### 4.2 Global Reach and Audience Insights

Beyond impressions,some insights like geography, demographics, and traffic source are only available on YouTube. While the geography and demographics data encompasses all Tech

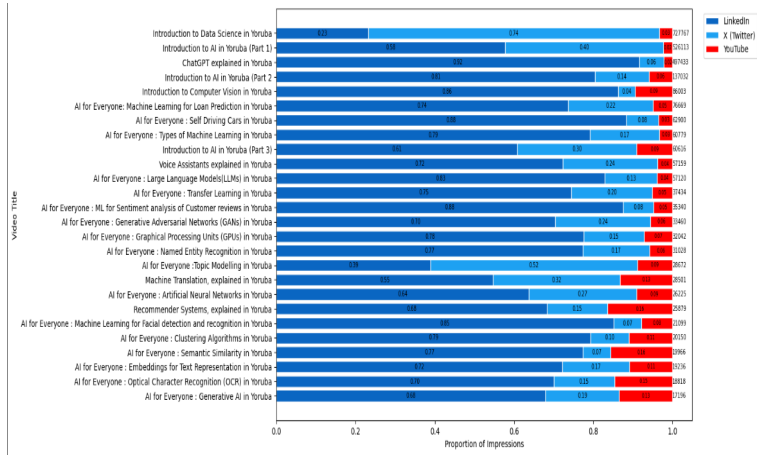

Figure 2: AI in Yoruba Video Impressions: Count and Proportion Across Platforms

in Yoruba videos published between May 2022 and February 2024, not just AI-focused ones, it provides valuable context for understanding how viewers interact with AI literacy content in low-resource languages. The videos have been viewed in 22 countries: Nigeria, United Kingdom, United States, Canada, United Arab Emirates, Germany, South Africa, Finland, India, Ghana, Malaysia, France, Ireland, Poland, Australia, Italy, Qatar, Estonia, Benin, Belgium, Côte d'Ivoire, and the Netherlands. This global reach demonstrates that interest in this type of content transcends the geographical areas associated with Yoruba language speakers.This highlights the value of including English subtitles to cater to a wider audience, including native speakers, first and later generation migrants, and non-speakers seeking technical information or language learning opportunities. With an average video length of 6 minutes, Figure 3 shows the view percentage distribution across age groups for Tech in Yoruba videos on YouTube. Viewers between 25 and 44 years old contributed the most (44%), with an average watch time of 1 minute, 41 seconds. The 35-44 age group followed at 36% (2 minutes, 6 seconds average watch time). Viewers aged 45-54 contributed 9.7% (2 minutes, 23 seconds), while younger viewers (18-24) contributed 5.8% (1 minute, 26 seconds). Older viewers (55-64 and 65+) contributed 2.67% and 1.73% respectively, with longer average watch times (2 minutes, 23 seconds and 2 minutes, 36 seconds) This suggests that the main audience for Tech in Yoruba videos lies between 25 and 44 years old, a potentially crucial demographic as they might have young children under 18, opening doors for engaging them with educational AI literacy content. While older viewers (>55 years) tend to watch for longer, they represent a smaller share. It is important to remember that this analysis is based solely on YouTube data, and distribution on other platforms might differ. Additionally, platform demographics and potential language barriers for younger viewers could contribute to the lower viewership percentage in the 18-24 age group.The average watch time suggests that the target total video length of around 5 minutes is likely the optimal choice for engaging viewers across age groups

4.3 TRAFFIC SOURCES AND ACQUISITION STRATEGIES

Figure 4 shows the number of views from AI in Yoruba videos on YouTube. Notably, over half (52.8%) originated from external sources, with channel pages (9.2%), suggested videos (8.7%), and browse features (8.1%) contributing significantly. Direct or unknown sources accounted for 7.8%, with YouTube search (5.4%), other YouTube features (2.5%), playlists (3.3%), notifications (2.0%), end screens (0.2%), and hashtag pages contributing less than 0.00015%. While limited to YouTube, this data offers valuable insights into effective acquisition channels. The high percentage of traffic from external sources highlights the potential of promoting content beyond YouTube. Interestingly, hashtag pages played a negligible role, suggesting the need for exploring alternative strategies for reaching viewers.

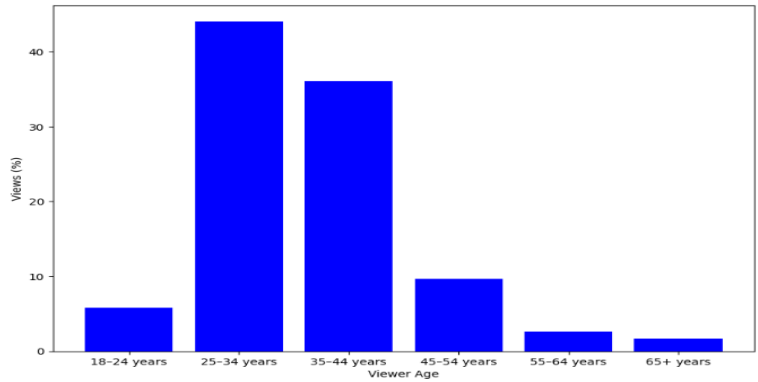

Figure 3: Percentage of Viewers vs Age Group for Tech in Yoruba YouTube videos

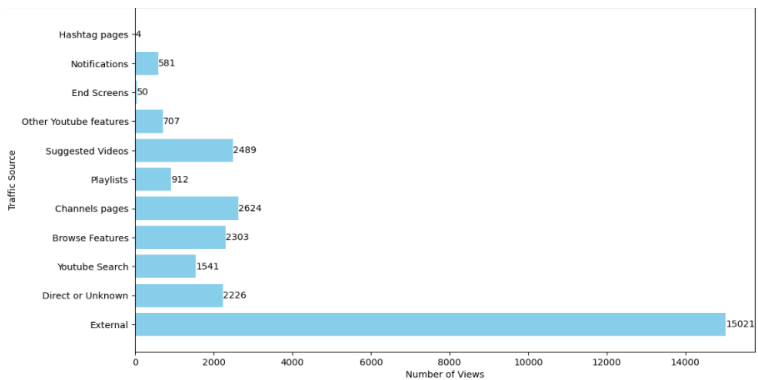

Figure 4: AI in Yoruba video views by Traffic Source(Youtube)

## 5 Conclusion

This work significantly contributes to AI literacy education in low-resource languages. By creating and distributing 26 AI in Yoruba videos, the project garnered a remarkable 2,744,637 impressions across channels, demonstrating not only the feasibility but also the potential impact of such initiatives. It presents a replicable and cost-effective methodology, documented in this paper, including the development of a 22-word vocabulary in Yoruba for key AI concepts. Furthermore, analysis of real data provided valuable insights into topics, demographics, geography, and acquisition sources across various channels. This data can inform the creation of targeted and engaging content for diverse audiences, like the high viewership among the 25-44 age group on YouTube.

Beyond its immediate impact on AI literacy in Yoruba, this project serves as a springboard for similar initiatives in other low-resource languages. It demonstrates the potential for AI literacy education to empower communities, preserve languages, and promote their continued use in the digital age. This work encourages researchers, educators, and language enthusiasts to build upon this foundation and create a future where everyone, regardless of their language background, has access to the knowledge and skills needed to thrive in an increasingly AI-driven world.